# Chapter 2 AI-aided Decision-Making in Education – Uncertainty, Explainability, and Human Responsibility

**Fabian Kieser*[,1], Paul Tschisgale*[,2], Peter Wulff[1]**

[1] Physics and Physics Education Research, Ludwigsburg University of Education, Ludwigsburg, Germany
[2] Department of Physics Education, Leibniz Institute for Science and Mathematics Education, Kiel, Germany
* Kieser and Tschisgale contributed equally and share first authorship.

**Abstract** As Artificial Intelligence (AI)-based systems have become increasingly integrated into various aspects of everyday life, including education, understanding how these systems make decisions is of paramount importance. AI holds significant potential to enhance educational practices through the facilitation of personalized and adaptive learning, automation of assessment procedures, and assistance with administrative tasks. However, the opacity of AI-based decision-making in education (especially with deep learning and generative AI models) can hamper decision-making, especially when outcomes appear unjustified or incorrect. In this chapter, we argue that three key principles—uncertainty, explainability, and human responsibility—are essential for the legitimate use of AI in educational settings. Acknowledging the inherent uncertainties in AI-based decisions, analogous to those in human decision-making, is fundamental. Explainability is critical for ensuring that the functioning and outcomes of AI applications are transparent and comprehensible to both educators and learners. Furthermore, human responsibility is crucial, particularly in human decision-making, which remains indispensable across all stages of AI development and deployment. By examining these three principles, this chapter provides a comprehensive guide for stakeholders seeking to integrate AI thoughtfully and effectively into the educational system.

# 1 Introduction

Inextricably linked to human autonomy is the necessity to understand decisions related to oneself and others, particularly in education-related high-stakes contexts such as examinations or university admissions (e.g., Miller, 2019). With the increasing sophistication and permeation of AI-based systems into almost all areas of everyday life, AI-aided decision-making processes have become more complex and, consequently, increasingly challenging to understand and explain for non-AI specialists, and even AI-experts (e.g., Osasona et al., 2024). Currently, educational settings undergo a kind of AI-incited transformation as AI-based systems are employed with the aim of providing personalized and adaptive learning, e.g. through learning platforms (e.g., Akgun & Greenhow, 2022). How decisions in such learning systems are made generally remains unknown to users, which creates dissatisfaction and reduces trust if decisions seem unjustified or factually wrong (Doshi-Velez & Kim, 2017). Consequently, the explainability of AI-aided decisions is a central aspect of legislative documents such as the General Data Protection Regulation (General Data Protection Regulation, 2018) and the Artificial Intelligence Act (Artificial Intelligence Act, 2024), which demand disclosing the decision-making processes of AI systems, including potential limitations and biases, to allow people affected by those decisions to understand and eventually challenge them.

How does this relate to everyday actions in the school context? Imagine a physics teacher who must evaluate his students' physics performance in an end-of-class examination. Assume that the exam can reliably and validly measure students' physics problem-solving abilities. Now, this teacher intends to shortcut his work by using a recently developed AI tool that could actually grade exams. While this is certainly a future perspective, since grading takes up a lot of teachers' valuable time (and sometimes gradings are unreliable due to stressors such as fatigue) and AI technologies are on the rise (see Kortemeyer et al., 2024), we argue that certain fundamental principles must be fulfilled for such a future perspective to become desirable: 1) an understanding of the inherent uncertainty of AI-aided decision-making must be established; 2) questions related to the explainability of the AI-based system must be reconciled to a degree that substantial evidence guarantees the reliability and validity of scores and inferences; and 3) human responsibility as well as accountability for making final decisions that might be consequential for students' future lives, must be conclusively considered.

In the upcoming sections, we will discuss the above principles for meaningful and legitimate use of AI-based systems in educational settings. Thus, this chapter serves as a comprehensive guide for stakeholders looking to effectively and thoughtfully integrate AI into the educational system. We begin by providing an overview of how AI-based systems are already widely used in education and educational decision-making. We then elaborate on the concept of *uncertainty* in AI-based decisions and its role in educational decision-making, and why understanding this concept is crucial for a meaningful adoption of AI systems. Subsequently, we address the *explainability* of AI-based decisions, offering insights into explainable AI methods, which represent one way of enhancing clarity with regard to understanding decision-making governed by AI. With regards to the third principle, we elucidate the *human responsibilities* associated with the utilization of AI-based systems in education, emphasizing that informed human decision-making is indispensable throughout all stages of AI development and implementation.

## 2 AI-based Applications in Educational Settings

Learning is a complex process that differs among individuals and depends on a wide variety of factors. Hence, a large number of observations – or data in general – are required to better understand it. These data is available in the form of test scores, written essays, and log files from learning platforms. Since educators' decision-making does not scale well with the increasing volume and complexity of this data, only some of it is typically used by educators for their decision-making, thereby neglecting potentially valuable diagnostic information. By leveraging the potential of AI, patterns and trends can be identified in vast amounts of educational data, thereby enhancing (or distracting!) an understanding of diverse learning processes and offering opportunities to improve educational outcomes by addressing students' individual needs. Currently, a plethora of AI-based systems are used for a wide variety of tasks in education. Relying on the (slightly adopted) taxonomy developed by Holmes & Tuomi (2022), we distinguish three types of AI-based systems: learner-focused AI, teacher-focused AI, and institution-focused AI.

*Learner-focused AI* refers to applications that support individuals' learning by automatically assessing their learning to provide individualized feedback and instruction, possibly even by adapting this feedback and instruction to learners' individual needs. For example, AI was employed to automatically assess text-based responses (Gao et al., 2024), concept maps (Bleckmann & Friege, 2023), and hand-drawn scientific models (Li et al., 2023, Zhai et al., 2022). All of these studies demonstrate the potential for automatically providing feedback and instruction (Steinert et al., 2023). If this feedback or instruction should also be adaptive, then AI-based systems require a so-called learner model (also called student model; Chrysafiadi & Virvou, 2013) that tracks a learner's knowledge, skills, progress, preferences, and possibly other things. These learner characteristics are then used to provide adaptive and individualized instruction and feedback, to adjust a task's difficulty, or to reframe a task's context. One specific example of learner-focused adaptive AI-based systems are Intelligent Tutoring Systems (ITS; Alkhatlan & Kalita, 2019; Mousavinasab et al., 2021). Such systems can determine instructional sequences and materials based on learners' characteristics that are stored in a learner model. More specifically, such systems leverage an inner and an outer loop to enhance adaptivity and effectiveness (VanLehn, 2006). The inner loop provides hints and feedback based on the learner's current performance in a specific task. The outer loop focuses on long-term learning by continuously updating the learner model based on the data generated within the inner loop, allowing for the planning of personalized learning paths (e.g., determining the next tasks to solve). Thus, decision-making within an ITS can be considered a complex and multi-layered process. Such ITS were shown to be particularly effective instructional tools (Kulik & Fletcher, 2016).

*Teacher-focused AI* aims to support teachers in lesson planning and classroom management. With regard to lesson planning, ChatGPT, as an AI-based web application, was used to develop instructional materials such as physics problems (Kieser & Wulff, 2024; Küchemann et al., 2023) and to reframe problems across academic domains (Einarsson et al., 2024). Moreover, use cases such as probing instructional materials (Kieser et al., 2023) and automatically transcribing and scoring exams (Kortemeyer et al., 2024) were suggested based on the specific capabilities of the proposed systems. AI also has the potential to transform classroom management, particularly in environments with high student-to-teacher ratios, which generally impedes individualized support. One notable benefit of AI-based systems in education is their ability to perform automatic and rapid formative and summative assessments (e.g., Zhai et al., 2020). Moreover, there are AI-based applications that allow real-time monitoring of learners' progress, such as dashboards (Ahmad et al., 2023). A dashboard is an interface displaying key metrics giving teachers either detailed information or overview data regarding the learning progress of the entire class or individual students, which can inform educational decision-making in classroom contexts. In this way, dashboards can highlight trends, pinpoint topics or tasks that students struggle with, and recommend tailored interventions, thus supporting informed decision-making in classroom settings.

*Institution-focused AI* aims to streamline and optimize administrative and operational processes while enhancing decision-making within educational institutions. A prominent application of institution-focused AI is in early warning systems that analyze vast amounts of data to identify students at risk of failing an exam or dropping out of school (e.g., Chung & Lee, 2019; Faria et al., 2017; Pace et al., 2024). These systems enable educational institutions to intervene proactively by providing targeted support to students who most need it. Another prominent use case is admission procedures for higher education and involved decision-making (Mengash, 2020). AI-based systems can scan and evaluate large volumes of applications quickly, thereby reducing the workload and associated costs. Using such AI-based technologies also comes with the idea of bypassing human biases (Holmes & Tuomi, 2022). However, this approach has faced criticism as it can inadvertently perpetuate existing biases at scale if the underlying data or algorithms are biased (Mehrabi et al., 2022).

The integration of AI-based systems in education undeniably holds significant potential to enhance educational decision-making, yet it also raises important considerations regarding the varying impacts these applications can have on students' lives, distinguishing between low- and high-stake scenarios. AI-based systems used in low-stakes scenarios, such as adaptive learning platforms and automated feedback systems, pose a limited risk to students' overall educational outcomes and future prospects. However, AI-based systems used in high-stakes scenarios, such as automatic scoring systems for high-stakes testing, early warning systems, and AI-driven admission procedures, can significantly influence a student's educational trajectory and future opportunities. In these high-stakes scenarios, it is essential that the decision-making processes of AI-based systems (or at least the crucial parts thereof) are explainable, ensuring that users can understand and trust the basis of these critical decisions. As AI-based systems become increasingly complex, hindering easy understanding of AI-based decisions, the demand for transparency and explainability increases. In March 2024, the European Parliament approved the Artificial Intelligence Act, which establishes obligations in the European Union for AI-based systems according to their potential risk (Artificial Intelligence Act, 2024). This risk-based approach categorizes AI-based systems into four risk levels: minimal, limited, high, and unacceptable. According to this Act, AI applications in education are mostly considered high risk since these AI systems "may determine the educational and professional course of a person's life and therefore affect that person's ability to secure a livelihood" (Artificial Intelligence Act, 2024). This categorization implies that all AI-based systems in education are subject to obligations regarding risk management, data governance, documentation, transparency, and human oversight. The associated bureaucratic expenses are substantial and may ultimately impede their overall usage. To ensure a continued use and development of widely employed low-stakes educational AI-based systems that facilitate and support individualized learning on a broad scale, we appeal for a more refined categorization of educational AI-based systems within the AI Act.

Overall, there exists a wide consensus that AI-based systems have the potential to enhance learning and teaching processes; however, they present novel challenges, particularly regarding AI-aided educational decision-making. When AI-aided decisions are made in high-stakes scenarios, recognizing the uncertainty inherent in AI-based decisions, as well as quantifying and communicating this uncertainty, becomes crucial. Additionally, aspects of transparency and explainability of AI systems are essential for high-stakes educational decision-making. Human oversight and responsibility remain central. Educators and administrators should be equipped to interpret AI recommendations, challenge potential biases, and make informed decisions based on AI outputs. Thus, considering uncertainty, explainability, and human responsibilities in AI-aided decision-making is of utmost importance to safeguard the interests of all stakeholders involved, and AI-based outputs should augment, rather than replace, human decision-making and agency (e.g., German Ethics Council, 2023).

## 3 Uncertainty

The certainty or uncertainty of a given AI-based decision – especially in high-stakes scenarios – is often overlooked. Uncertainty is often confused with other metrics such as predictive accuracy or precision. To clarify these concepts, consider a hypothetical AI-based early warning system that detects whether students will pass their final physics exam which involves solving multiple physics problems. Such a system may be particularly valuable to identify struggling students early and offer additional support measures. The AI-based system was trained on recent data that reflects students' problem-solving abilities and includes information on whether they passed or failed the final physics exam. Moreover, the AI-based system was trained to maximize predictive accuracy, i.e., to maximize the proportion of students who were correctly predicted to pass or fail. The system was tested on a hold-out test data set of 1000 students. Table 1 presents the hypothetical testing results.

**Table 1.** Contingency table illustrating the performance of the hypothetical AI-based early warning system on the test data set.

| | Student failed final exam | Student passed final exam |
|---|---|---|
| Student was predicted to fail | 30 (True Positives) | 50 (False Positives) |
| Student was predicted to pass | 20 (False Negatives) | 900 (True Negatives ) |

The predictive accuracy of this hypothetical early warning system, defined as the fraction of correctly categorized students, is (30+900)/1000 = 93%. At first glance, this appears promising. However, a closer examination reveals that predictive accuracy is not the most meaningful measure for this decision problem. Instead, the quality of the classification of students who failed their final physics exam is more important. In this example, only 30 of the 50 students who failed were correctly predicted to fail. This fraction is called sensitivity (or recall) and is only 30/50 = 60%, which is only slightly better than a (fair) coin toss. In other words, this early warning system fails to identify 40% of students who failed their final physics exam, rendering it rather insufficient and unethical if used in practice. After all, 20 students would not have received a warning, whereas 30 other students were “privileged” to have received the warning.

Now, consider a scenario where a teacher receives a notification from the early warning system indicating that a specific student is at risk of failing the final exam. How certain is this AI-based decision? Looking at Table 1, we are interested in the fraction of students who actually failed among all those predicted to fail, which was 30/(30+50) = 37.5%. This value, known as precision, suggests that *on average* only 37.5% of the students predicted to fail actually failed the final exam. However, a teacher who must make a decision about raising additional resources to support specific students (e.g., sending them to summer school) is not interested in *average*s but in the uncertainty of that specific AI-based decision. Precision and other metrics based on the values in Table 1 relate to multiple decisions about multiple individuals; however, we need to understand the uncertainty related to a single decision impacting one specific individual.

The need to quantify the uncertainty of AI-based decisions becomes particularly clear in high-stakes scenarios. Each field in Table 1 is associated with a different impact, with the highest (and worst) impact linked to False Negatives – students at risk of failing who are not detected by the early warning system. The fraction of students who actually failed among those predicted to pass was 20/(20+900) ≈ 2.2%, indicating that this happens rarely on average. However, regardless of this average rate, each prediction or decision carries a specific uncertainty because the AI-based system is typically not equally confident in each prediction. In line with Christian (2021), we argue that AI-based systems should generally avoid making high-impact decisions if significant uncertainty is involved. Therefore, if the hypothetical early warning system's prediction of a student passing the final exam is highly uncertain, this should prompt the teacher to examine that student more closely in order to make the best decision (under uncertainty) for the student.

Uncertainty quantification in AI-based decision-making is an active field of research (Abdar et al., 2021; Hariri et al., 2019). It is argued that quantifying the uncertainty of AI-based decisions is a key component of every AI-based system, particularly in high-stakes scenarios (Hüllermeier & Waegeman, 2021). While the necessity for uncertainty quantification in AI-aided medical decision-making is self-evident (Begoli et al., 2019), AI-aided educational decisions can also significantly impact students' futures. Thus, we urge developers of educational AI systems to implement uncertainty quantification procedures (see e.g., Abdar et al., 2021; Hüllermeier & Waegeman, 2021), particularly when employed in high-stakes scenarios.

## 4 Explainability

As should be clear right now, AI-based systems offer new potential in various settings including educational contexts. However, how decisions of educational AI systems are generally made remains unknown to users which can create dissatisfaction and reduce trust if these decisions seem to be unjustified or obviously wrong. According to Article 15 of the EU's General Data Protection Regulation (General Data Protection Regulation, 2018), humans have the right to obtain an explanation of the underlying logic of automated decision-making that impacts them. Moreover, the recently adopted AI Act (Artificial Intelligence Act, 2024) proposes rules and obligations regarding the explainability of AI-based systems, particularly for the education sector. An AI-related definition of explainability, following Doshi-Velez and Kim (2017)[1], refers to the ability to explain or represent the behavior of AI-based systems in a way that humans can understand. However, the inner workings of decision-making in AI-based systems are often unclear. Consequently, an entire branch of research focusing on making the decision-making processes of AI comprehensible and explainable emerged (Xu et al., 2019). The large number of application scenarios for AI methods has led to terminological inconsistencies around the concept of explainability. For example, the terms *transparent*, *interpretable,* and *explainable* are often used interchangeably (Graziani et al., 2023). Lipton (2018) pointed out that the inconsistent use of these terms can lead to misunderstandings and therefore developed a taxonomy that places explainability at the top and subordinates the sub-concepts of transparency and interpretability (see Figure 1)[2].

[1] [Added after publication.] Doshi-Velez and Kim (2017) give this definition for *interpretability*, not *explainability*. The mislabeling resulted from an oversight by the authors; see footnote 2 for how the terms are used in this chapter.

*Transparency* of an AI-based system primarily refers to the disclosure of its technical properties. Lipton also distinguishes between three aspects of transparency (see Figure 1). The first aspect is *simulatability.* In easy terms, simulatability suggests that a user or developer can effectively "think along" with the AI-based system by understanding and following each step that the model takes in processing its input data and arriving at a decision. For instance, consider the hypothetical early warning system introduced in the last section, which detects whether students will pass their final physics exam based on their assessed physics problem-solving ability. For simplicity, we suppose that this ability is composed of two factors that were measured: physics knowledge and meta-knowledge of how to solve physics problems. Simulatability would mean that a user of the early warning system can manually trace and understand the weight and impact of both physics knowledge and meta-knowledge on the final decision (i.e., predicted to pass or fail the final exam) given by the AI-based system. *Decomposability* is another aspect of the transparency of AI-based systems. This basically means that each component of the system – ranging from inputs and its inner structure (including model parameters and performed computations) to outputs – allows for an intuitive and comprehensible explanation (Lipton, 2018). To express this more succinctly, if each component of the model can be easily explained, then the model is more decomposable and, consequently, more transparent. *Algorithmic transparency* addresses the specific learning algorithm used to train an AI-based system. This refers to the extent to which a learning algorithm's inner workings and decision-making processes can be understood by humans (Lipton, 2018).

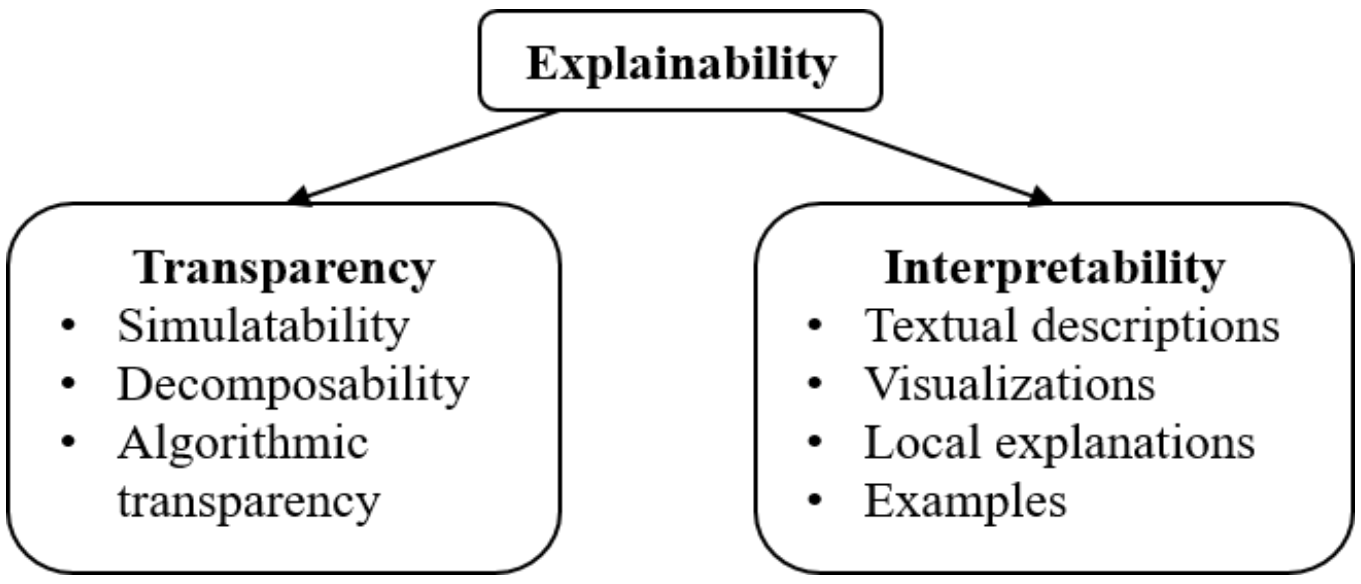

**Figure 1.** Taxonomy of the terms explainability, transparency, and interpretability developed by Lipton (2018).

[2] [Added after publication.] Contrary to this statement and to Figure 1, Lipton (2018) uses *interpretability* as the overarching concept, distinguishing between *transparency* and *post-hoc explanations*. In this chapter, *explainability* broadly corresponds to Lipton's *interpretability*, whereas *interpretability* is used in a narrower sense, referring primarily to post-hoc explanations. Accordingly, "ad-hoc interpretations" in this section should read "post-hoc interpretations". This discrepancy resulted from an oversight by the authors, not from an intentional departure from Lipton's conceptual framework.

In summary, the transparency of an AI-based system entails clear disclosure of its technical properties, emphasizing aspects such as simulatability, which allows users to mentally follow and understand each step of the decision-making process; decomposability, which ensures that every component of the model is intuitively explainable; and algorithmic transparency, which focuses on the comprehensibility of the learning algorithm's inner workings and decision-making processes. This clarifies that good documentation plays a crucial role in achieving transparency in AI-based systems by providing detailed and understandable descriptions not only for developers but also for users and other stakeholders. A good documentation should clearly outline what training data were used, explain how the training algorithm works, detail the decision-making process of the system by elucidating the role and function of each component of the model, and describe how the outputs are interpreted.

In addition to transparency, the *interpretability* of AI-based systems hinges on contextual nuances to elucidate the processes and rationales behind AI-based systems. There exist explainable AI techniques that aim to provide end users (who could be teachers or students) with useful information on how and why an AI-based system produces certain outputs or makes specific decisions, thus providing ad-hoc interpretations of an AI-based system's behavior (e.g., Khosravi et al., 2022; Samek et al., 2019). According to Lipton (2018), one aspect of interpretability is *textual explanations* that can accompany AI-based decisions, similar to how real humans sometimes justify their decision-making verbally (with frailties such as reconstruction errors, distorted memory, etc.). *Visualizations* can also contribute to the better interpretability of AI-based systems. AI-based systems often work with high-dimensional mathematical representations (e.g., Mikolov et al., 2013). For example, to make complex data easier for stakeholders to understand, it is helpful to transform high-dimensional data into easy-to-visualize two-dimensional data. These two-dimensional data can be easily visualized, which would enable easier interpretations (for educational examples, see Martin et al., 2023; and Tschisgale et al., 2023). Lipton (2018) also suggests using *local explanations* to improve interpretability. Local explanations seek to elucidate how an AI-based system arrives at a specific prediction by either fitting a simpler, local model around a particular decision, or by altering the values of input variables to assess how these changes impact a model's output (Mittelstadt et al., 2019). This allows the relevant features to be identified. For example, feature assignment techniques that assign importance values to the inputs of complex AI models' inputs (which are also referred to as features) exist, thereby quantifying the influence of the inputs on the models' outputs (e.g., Lundberg & Lee, 2017). Finally, Lipton (2018) argues that *examples* can, to some extent, explain AI-based systems' decisions and hence contribute to interpretability. Lipton (2018) specifically refers to the prominent *word2vec* algorithm (Mikolov et al., 2013), which represents words, or more precisely their meaning, by vectors (i.e., points) in a high-dimensional vector space. As illustrated by Mikolov et al. (2013), representing words as vectors allows constructing (approximate) "equations" such as

$$\text{Paris} - \text{France} + \text{Italy} \approx \text{Rome}.$$

Using the *word2vec* algorithm, we wanted to find the word that is the best solution to the physics-related "equation"

$$\text{physics} + \text{stars} \approx ?$$

We found that the best solution to this "equation" is the word *star*. However, the second-best solution was the word *astrophysics*, which was our initial expectation. Overall, analyzing examples of such representational vectors contributed to the interpretability of the *word2vec* algorithm.

In summary, transparency generally refers to the disclosure of the technical properties of AI-based systems, whereas explainability goes beyond transparency by also aiming to elucidate the processes and rationales behind AI-based decisions, focusing on the interpretability of these systems within contextual nuances.

The aforementioned notions of explainability have proven effective for a wide range of AI-based systems, providing insights into their decision-making processes and enhancing user trust. However, when it comes to increasingly employed Large Language Models (LLMs, such as generative pretrained transformers, GPT, as in the conversational AI ChatGPT), these traditional methods face significant challenges. An LLM is a type of AI system that processes and generates human-like text based on patterns and information learned from vast datasets. The sheer complexity and scale of LLMs, characterized by their vast parameter spaces and intricate data processing mechanisms, render standard explainability techniques insufficient (Zhao et al., 2024). This disparity underscores the necessity of innovative approaches designed to elucidate the unique operational dynamics of LLMs, ensuring that their decisions remain explainable in practical applications. Hence, there are also explainable methods specifically designed for LLMs that intend to explain their outputs. Although traditional feature assignment techniques (e.g., Lundberg & Lee, 2017; Sundararajan et al., 2017) could theoretically be applied to LLMs, their practicality is limited by the immense computational resources required due to the billions of parameters in LLMs. Consequently, these traditional explainable AI techniques are less suitable for LLMs. A comprehensive overview of explainable AI methods specifically designed for LLMs is detailed in Zhao et al. (2024), including various techniques for prompting strategies that help explain LLM's decision-making. These techniques are becoming increasingly important in the educational sector, where LLMs are used for a range of applications such as automated feedback tools (e.g., Wan & Chen, 2024).

Scientific literature frequently emphasizes the explainability of algorithms as a crucial prerequisite for trust in AI-based systems (e.g., Ribeiro et al., 2016). This is particularly true in educational contexts, where understanding how AI-based systems arrive at decisions is paramount. Improved explainability not only fosters greater acceptance but also encourages broader adoption of AI-based systems.

# 5 Human Responsibility

Humans play a decisive role in many stages within the development and deployment phases of educational AI-based systems (Holmes & Tuomi, 2022; Ninaus & Sailer, 2022). We start by discussing the responsibilities of developers[3] of such systems and then clarify the responsibilities of users (e.g., educators and learners).

## *5.1 Responsibilities of Developers of AI-based Systems in Education*

Developers of AI-based systems employed in educational settings have great responsibilities as their systems will always directly or indirectly impact students' learning and thus ultimately have an influence on students' future lives.

The majority of AI-based systems are based on data-driven AI models (such as LLMs), i.e., these models are trained on enormous amounts of data. In addition, AI-based educational systems that promise to be adaptive (e.g., Duolingo; see Munday, 2017) also have to build a learner model, i.e., data of the user is collected to provide tailored instruction or feedback. Hence, *data governance* is a central responsibility of developers. This involves aspects of data privacy and data protection, e.g., preventing the leakage of sensitive data (Weidinger et al., 2021). In the industrial sector, such regulations are typically established by law, whereas in research, data protection officers and similar bodies ensure compliance with the data protection standards. Ensuring data privacy and protection is important; however, developers must first decide what data must be collected. Here, the principle of data minimization is central, which means collecting only the amount of data that is strictly necessary for the specific purpose of an AI-based system, avoiding excessive or unnecessary data gathering. For our hypothetical AI-based assessment tool for problem-solving abilities, it might be unnecessary to collect data on students' gender or spelling mistakes as this data would not be useful for providing feedback[4]. In addition to ensuring data minimization, it is crucial to ensure the quality of the data used to train the AI-based systems involved. This includes ensuring the representativeness of the data, both in terms of the underlying population and changes over time. For example, if we plan to develop an assessment tool for problem-solving abilities for the whole range of students in high school, only using training data from students in the senior year would not make a representative training sample. Moreover, the average characteristics and abilities of student cohorts are expected to change over time. This is why older training data might be discarded one day and new, more representative data needs to be collected, as the problem-solving abilities of, for example, high-school students in 2025 may differ in distribution from those of high-school students in 2030. Thus, training data needs to be regularly reviewed and updated to reflect the changing context of the world, schools and individuals. These data must, of course, be reliable and valid, ensuring they are almost free of measurement errors as well as correctly labeled, and appropriately represents the aspects of the world we are actually interested in. For example, if we are interested in students' physics problem-solving abilities but only gather data on their final results in physics problems, these data would not be valid for representing problem-solving abilities, as the entire process leading to the final result incorporates important aspects of problem solving (e.g., Docktor et al., 2016). Overall, ensuring a high quality of data is a first step in mitigating bias in AI-based decision-making as bias often enters AI-based systems through their training data (Mehrabi et al., 2022). At the same time, data protection and privacy must be ensured to gain users' trust and confidence in the use of their data.

[3] By developers, we mean all individuals involved in the development process of an AI-based system, including researchers, engineers, designers, and other related professionals.

[4] Naturally, collecting such data could be valuable for research purposes, provided that clear hypotheses regarding the data are established in advance. However, it is essential to inform users in general, and students in particular, that such data is being collected.

While developers must decide which data to collect, they must also make a *decision on the employed AI technique*. Ninaus and Sailer (2022) argue that the decision for or against a specific AI technique "should not only be data-driven but also informed by theory and determined by the overall goal." For example, solely data-driven decision-making tests a range of AI techniques and finally chooses the one with the best predictive accuracy (or any other performance metric) on the available data. Instead, developers can just use multiple AI techniques and pool their results. This would also allow quantifying the uncertainty of AI-based decisions, as we can compare the individual decisions of all used techniques (Abdar et al., 2021). However, there is more to think about: while one AI system may perform slightly worse than another in terms of, for example, predictive accuracy, it might be worth considering the worse system if its results are better explainable. It is important to select the appropriate criterion to "optimize" for in accordance with the overall goal. For example, if the goal of an AI-based system is to determine students who are at risk of failing the final exam, then the underlying AI model should not be selected based on the best predictive accuracy (which overemphasizes the larger proportion of correctly identified students who did not fail), but rather selected based on the recall (the proportion of students at risk who are correctly identified). When discussing the choice of which performance metric to "optimize" for, we need to take Goodhart's Law into account, which basically states: "When a measure becomes a target, it ceases to be a good measure" (Strathern, 1997). Thomas and Uminsky (2020) specifically argue that "optimizing" for a single metric can result in outcomes that are far from optimal. As possible solutions, for example, they propose considering multiple relevant metrics at once for deciding on a specific model as well as including stakeholders in the evaluation of AI-based systems.

Developers must also ensure *explainability* of their AI systems outputs. As discussed in Section 4, this involves transparency of the system (what are the technical properties of the system?) and the interpretability of the system's output (why do specific inputs lead to specific outcomes?). However, this is not trivial, as there is generally a trade-off between an AI model's performance (e.g., predictive accuracy) and the model's degree of explainability. Simpler models (e.g., linear regression models) generally provide poor predictions but are highly explainable, whereas complex models (e.g., deep neural networks) generally provide better predictions but are far less explainable (Molnar, 2022).

Having understood the specific patterns learned by an AI-based system through an explainable AI technique, developers can identify unintended biases, risks and other areas for improvement. Explainability can therefore help improve AI-based system (e.g., Strobelt et al., 2018). In particular, trying to make an AI system more explainable can help identify issues related to fairness in algorithmic decision-making (Mitchell et al., 2021). Fairness is crucial and requires that decisions are not based on unfair or discriminatory criteria (Mehrabi et al., 2022). For example, admission to university should not depend on gender to ensure fair and equal participation. AI-based systems have occasionally been found to discriminate groups of people by reproducing human stereotypes or social prejudices by creating harmful associations between certain characteristics and social identities (Weidinger et al., 2021). A well-known example involves prompting DALL-E (an AI-based system developed by OpenAI that generates images from textual descriptions) to "draw a picture of a scientist" leading to the result shown in Figure 2. DALL-E reproduced the stereotypical white male researcher in a white coat in some indoor laboratory (e.g., Ferguson & Lezotte, 2020).

Such biases, even if less obvious, can be present in any AI system trained on human data. Take, for example, the hypothetical AI-based problem-solving assessment system described earlier. Let us assume that this system used the previously discussed *word2vec* word representations (Mikolov et al., 2013). An analysis of selected physics terms and their similarity in vector space shows that some physics terms are distorted. The results of this analysis are illustrated in Figure 3. The word vectors corresponding to the words on the vertical axis capture some semantic information (as with the “*physics* + *stars*” example above). We investigated the similarities between specific physics terms and the two target words *“electricity”* and *“mechanics”*. These similarities are shown in Figure 3. Red (negative) indicates a closer proximity of the word vectors to “electricity”, while blue (positive) indicates a closer proximity to “mechanics”[5]. Observing Figure 3, it becomes clear that electricity-related physics terms (e.g., “voltage” or “electrons”) are generally more associated with the term “electricity”. However, words such as "experiments" or "field" occur in the physics context of both mechanics and electricity (e.g., gravitational field and electric field), but the *word2vec*-model assigns them more to the context of mechanics. These examples contribute to an understanding that biases (in the sense of systematic deviations) can occur due to the large datasets on which these AI-based systems are trained and the context-dependent nature of word meanings in domains such as physics (Wulff, 2024). Hence, by improving the explainability of AI-based systems' decision-making, developers can identify such biases, enabling them to take targeted measures to rectify them.

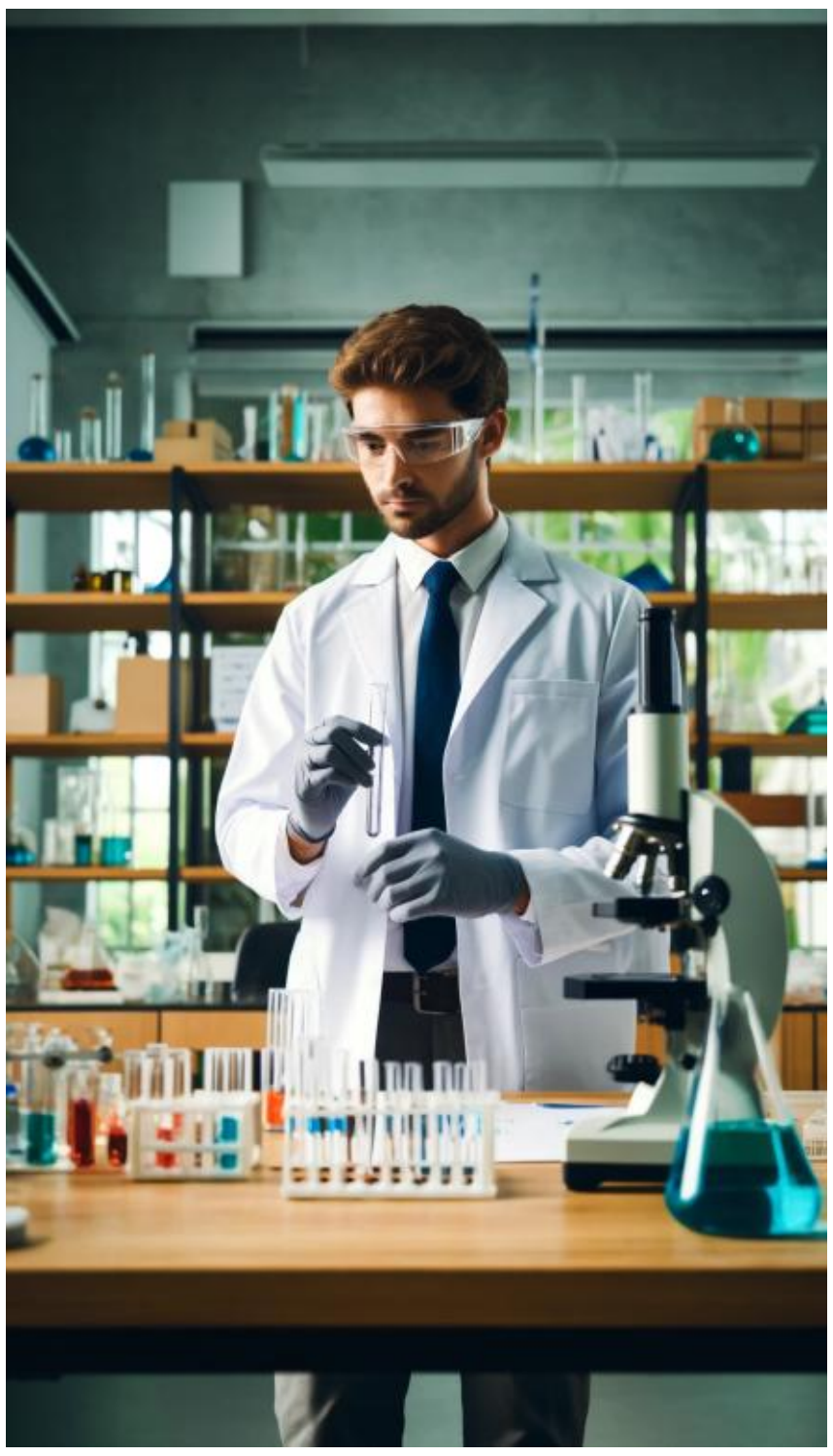

**Figure 2.** Output of DALL-E when prompted to "draw a picture of a scientist".

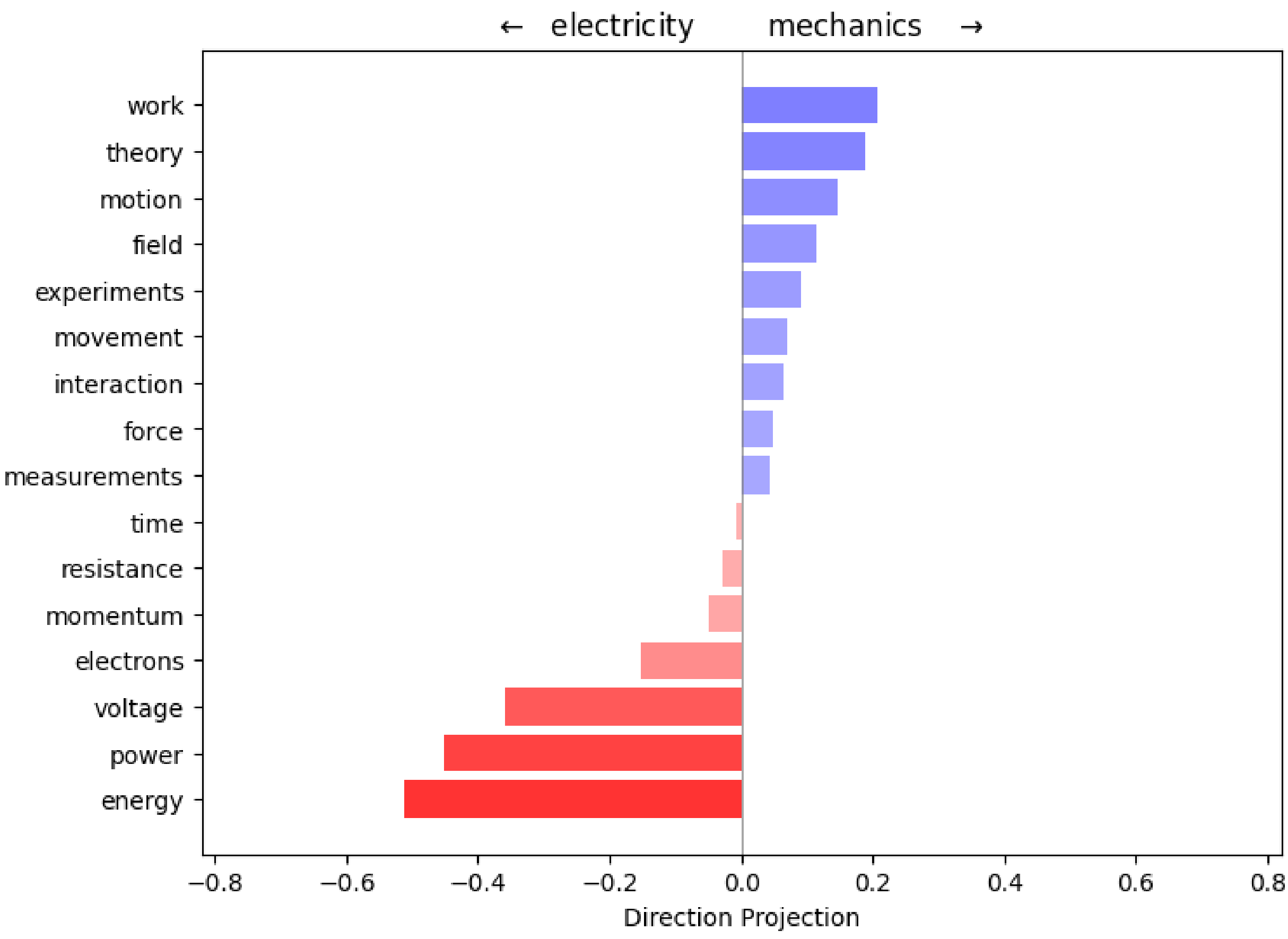

**Figure 3.** Analysis of the similarities of some physics terms in the *word2vec* model.

[5] We used the model *word2vev-google-news-30*. The distance metric was calculated according to the Word Embedding Association Test (WEAT; Caliskan et al., 2017).

Ensuring *students' (and humans') agency* (here, the capacity to act purposefully in institutional settings) is particularly important when using educational AI-based systems. As Christian (2021) argues, this should include the fact that students have access to and the ability to modify their respective student models inferred by the AI application. In doing so, students are empowered to assert that the model does not accurately represent them or that it aligns with who they aspire to be.

## *5.2 Responsibilities of Users of AI-based Systems in Education*

Responsibilities regarding AI-based educational systems do not rely solely on their developers. Educators and learners (e.g., teachers and students), as primary users, also bear numerous responsibilities. Understanding these responsibilities is crucial to ensure the effective and beneficial integration of AI-based systems into the educational landscape.

Both educators and learners must exhibit openness, flexibility, and adaptability towards AI-enhanced teaching and learning. Embracing AI systems requires a willingness to explore new methods and tools (Wang et al., 2020). Educators need to adapt their traditional educational practices to incorporate these innovations, while learners should be open to embracing these technologies. A fundamental understanding of AI is essential for both educators and learners to comprehend and evaluate AI-based decisions (Markauskaite et al., 2022). This has led to the emergence of the concept of *AI literacy* (Casal-Otero et al., 2023), which refers to the set of competencies that enables individuals to understand, use, evaluate, and ethically engage with AI-based systems, fostering effective collaboration with AI (Ng et al., 2021).

Continuous improvement and refinement of AI-based systems relies on reliable feedback from educators and learners regarding the usability and effectiveness of these systems. This feedback enables to adapt these systems to the evolving needs of educational environments. Since educators typically decide on whether and how to employ an AI-based system in their teaching, they are responsible for making this decision with care. Making an informed decision requires a comprehensive understanding of the capabilities and limitations of AI-based systems (Luckin et al., 2022). For effective integration into teaching practices, educators should only use AI-based systems within the defined area of application, i.e., for specific tasks and within the specific contexts for which they were trained. For example, an AI-based learning system trained using only data from high-achieving students should not be employed in low-achieving classes.

If the consequences of decisions hold significant weight which is characteristic of high-stakes situations, then humans (should) keep the responsibility for final decision-making. As of now, (trained) humans are better at focusing on relevant information and integrating it in ways that do minimal harm to the people about which decisions are made about. For example, an AI-based system may analyze students' academic performance, interests, and career goals to recommend possible courses to choose from. However, students ultimately make the final decision, the AI-generated decision only serves as guidance to potentially augment human decision-making. Similarly, learning platforms can plan personalized learning paths for students; however, teachers should review these plans and potentially adjust them.

Data privacy and security issues are also of paramount importance. Educators must be vigilant about safeguarding learner data and ensuring compliance with privacy regulations. This responsibility extends to understanding the privacy policies of the AI systems they use and taking proactive steps to protect sensitive information. Adequate training opportunities should be provided to educators to use AI systems effectively and responsibly (Lee & Perret, 2022). This training should be continuous to keep pace with the latest developments and best practices in AI within education (Pokrivcakova, 2019). Additionally, building a community of practices in which educators can share best practices and experiences with AI in education can foster a collaborative environment that enhances overall effectiveness (e.g., Kasneci et al., 2023).

Teachers should regard AI as a tool to enhance, rather than replace, traditional teaching methods. AI can enable personalized learning experiences but should complement, rather than substitute, human teaching (e.g., Molenaar, 2022). Educators also have the responsibility of fostering AI literacy among their learners early on (Druga et al., 2019), preparing them to navigate and critically assess a digital and AI-driven world. AI literacy includes a basic understanding of AI and its strengths and limitations. Learners require AI literacy to discern between useful outputs generated by AI-based systems and those that are biased or inaccurate, thus avoiding "unreflected acceptance" of AI-generated outputs (Krupp et al., 2023).

Protecting one's own personal information is another key responsibility of learners, and not just educators. Learners should be mindful of the personal data they share with AI-based systems and understand the importance of data privacy in protecting themselves from potential misuse. It has been widely shown that LLMs can incorporate and output (with specific instruction) personally identifiable information (Chen et al., 2023). Fair use practices are also important. Learners must learn to differentiate between ethical and unethical uses of AI, such as presenting AI-generated content as their own. Maintaining open lines of communication with educators regarding their experiences with AI-based systems is crucial. By discussing both the positive and negative aspects, learners can help educators better understand how these systems impact learning. This feedback can then be used to improve and refine AI-based educational systems.

The integration of AI into education brings a shared set of responsibilities to both educators and learners. By fostering openness, ensuring proper training, safeguarding data privacy, and promoting ethical use, educational communities can maximize the benefits of AI while mitigating potential risks. Additionally, it should be clear right now that human decision-making is indispensable across all stages of AI development and deployment.

## 6 Conclusion

AI has the potential to enhance educational decision-making processes in learner-focused, teacher-focused, and institution-focused AI applications. However, it is important to recognize that AI-based systems, despite their sophisticated capabilities, are not flawless and can make errors that lead to incorrect decisions in educational settings, potentially having serious consequences for learners.

It was posited that three key principles—uncertainty, explainability, and human responsibility—are essential for the legitimate utilization of AI in educational settings. Recognizing the inherent uncertainties in AI-based decisions, analogous to those in human decision-making processes, is of paramount importance. Explainability is vital to ensure that the operations and outcomes of AI-based systems are transparent and interpretable to educators and students, fostering trust and enabling effective educational strategies. Furthermore, human responsibility is paramount; educators must remain the primary decision-makers and be accountable for how AI insights are used to influence educational outcomes.

The successful integration of AI and human intelligence in education can lead to more efficient processes and improved learning outcomes: AI handles data-intensive tasks, while educators apply their pedagogical expertise, intuition, and ethical considerations to make informed educational decisions. Achieving this requires continuous awareness of the strengths and limitations of both AI and human contributions, along with the ongoing monitoring and adaptation of their collaborative efforts. This balanced approach is crucial for leveraging the full potential of AI in education while minimizing its risks, ensuring that AI enhances rather than detracts from traditional educational practices.

## Author Contributions Statement

Fabian Kieser and Paul Tschisgale contributed equally to both the creation of the initial draft and the subsequent editing of the manuscript. Peter Wulff reviewed and edited different versions of the manuscript.

## Language editing

In the process of refining the manuscript, language editing was conducted with the assistance of ChatGPT, a language model developed by OpenAI. ChatGPT was employed to enhance the clarity, coherence, and style of the text while maintaining the integrity of the original content.